# TractSeg - Fast and accurate white matter tract segmentation

Jakob Wasserthal [a,b], Peter Neher [a], Klaus H. Maier-Hein [a, c]

[a] Division of Medical Image Computing (MIC), German Cancer Research Center (DKFZ), Heidelberg, Germany
[b] Medical Faculty Heidelberg, University of Heidelberg, Heidelberg, Germany
[c] Section for Automated Image Analysis, Heidelberg University Hospital, Heidelberg, Germany
{j.wasserthal, p.neher, k.maier-hein}@dkfz.de

**Abstract.** The individual course of white matter fiber tracts is an important factor for analysis of white matter characteristics in healthy and diseased brains. Diffusion-weighted MRI tractography in combination with region-based or clustering-based selection of streamlines is a unique combination of tools which enables the in-vivo delineation and analysis of anatomically well-known tracts. This, however, currently requires complex, computationally intensive processing pipelines which take a lot of time to set up. TractSeg is a novel convolutional neural network-based approach that directly segments tracts in the field of fiber orientation distribution function (fODF) peaks without using tractography, image registration or parcellation. We demonstrate that the proposed approach is much faster than existing methods while providing unprecedented accuracy, using a population of 105 subjects from the Human Connectome Project. We also show initial evidence that TractSeg is able to generalize to differently acquired data sets for most of the bundles. The code and data are openly available at https://github.com/MIC-DKFZ/TractSeg/ and https://doi.org/10.5281/zenodo.1088277, respectively.

## 1. Introduction

White matter fiber tract segmentation enables detailed analysis of individual white matter tracts. It helps to characterise the healthy brain and identify areas containing abnormal morphology in diseased brains (Bells et al., 2011, Yendiki et al., 2011, Yeatman et al., 2012, Dayan et al., 2016). Such tractometry analyses are inherently dependent on the quality of the underlying tract segmentation, and thus require accurate delineation.

Currently, the most commonly employed approach for segmenting white matter tracts is *virtual dissection*: streamlines that correspond to anatomically well-defined tracts are manually extracted from a tractogram using combinations of inclusion and exclusion of regions-of-interest (ROIs) (Wang et al., 2007, Stieltjes et al., 2013, Thiebaut de Schotten et al., 2011). This is a very time-consuming process that has limited reproducibility due to the subjectivity in the human interaction. Therefore, a variety of automatic approaches for white matter segmentation have been proposed. They can be grouped into three categories: *ROI-based segmentation*, *clustering-based segmentation* and *direct segmentation*.

*ROI-based segmentation* approaches typically use information from a common atlas space that is registered to the subject in order to extract ROIs in the cortex (parcellation) or in the white matter. Streamlines are then filtered according to their spatial relation to these ROIs (Cook et al., 2005). Yendiki et al. (2011) used cortical and white matter tract atlases (binary maps) to specify the morphology of tracts. Wassermann et al. (2016) used a Freesurfer parcellation to characterize different tracts. Yeatman et al. (2012) used ROIs and a probabilistic white matter tract atlas (probability maps).

*Clustering-based segmentation* approaches group streamlines into anatomically (Siless et al., 2018) or spatially coherent clusters. These clusters are either manually (O'Donnell et al., 2016) or automatically (e.g. using prior knowledge in form of streamline atlases) assigned to anatomically meaningful fiber tracts (Jin et al., 2014). Garyfallidis et al. proposed an alternative approach wherein the streamline atlas can be directly incorporated into the clustering process (Garyfallidis et al., 2017). Clayden et al. and Labra et al. also used streamline atlases in combination with a tract similarity measure to segment tracts (Clayden et al., 2007, Clayden et al., 2009, Labra et al., 2017).

ROI- and clustering-based methods involve whole series of processing steps for atlas registration, tractography, parcellation or clustering. The resulting pipelines are rather complex, computationally expensive and tedious to fine-tune. For example, most methods base their inter-subject consistency on a registration between subject and atlas. This in itself is a non trivial task that requires quality checking and that might introduce subtle errors which are then propagated through all further processing steps.

*Direct segmentation* approaches circumvent the intermediate generation of streamlines, thus simplifying the processing chain by directly producing complete tract segmentations from the input images. A variety of direct methods has already been presented, employing techniques such as template matching (Eckstein et al., 2009), Markov Random Field optimization (Bazin et al., 2011), geometric flow-based segmentation (Guo et al., 2008, Jonasson et al., 2005), surface evolution (Lenglet et al., 2006, Descoteaux et al., 2009) and k-NN based classification (Ratnarajah et al., 2014). However, the segmentation quality currently reached by these approaches is limited, which in turn has forced recent research to focus on ROI- and clustering-based approaches.

We propose TractSeg as a novel approach to direct white matter tract segmentation that provides complete and accurate segmentations, yet is easy to set up, fast to run and does not require registration, parcellation, tractography or clustering. This is achieved via a fully convolutional neural network (FCNN) that directly segments white matter tracts in fields of fiber orientation distribution function (fODF) peaks. We used a semi-automatic approach to segment reference segmentations of 72 anatomically well-describe tracts in a cohort of 105 subjects selected from the Human Connectome Project (HCP) (Van Essen et al., 2013) to train and evaluate our approach. We compared our model to six other state-of-the-art tract segmentation methods.

# 2. Materials and Methods

An encoder-decoder FCNN that processes the data and generates tract probability maps form the core of our method. Figure 1 gives an overview of the proposed segmentation workflow.

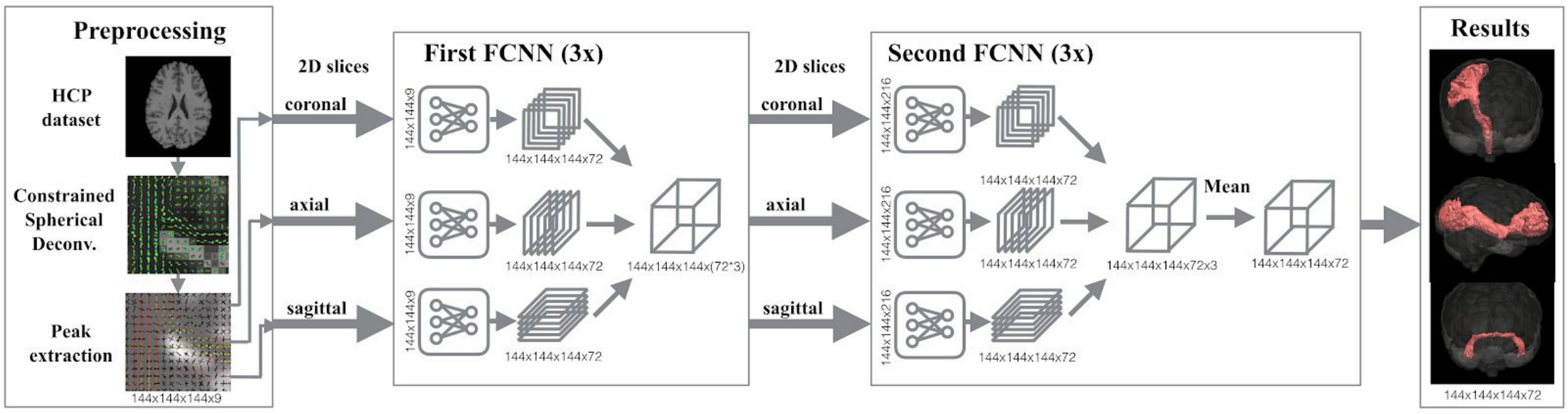

Figure 1: TractSeg segmentation pipeline. Constrained Spherical Deconvolution is used to extract the three dominant diffusion directions in each voxel. A 2D encoder-decoder FCNN then produces one tract probability image for each orientation (coronal, axial, sagittal) and for each tract. The tract probability images from the three orientations are then concatenated in the channel dimension resulting in a 3D image with 216 channels. This is used as input for a second FCNN which again runs three times. The three outputs per tract from the second FCNN are merged using the *Mean* to generate the final segmentation. The final segmentation is a 72-channel image, wherein each channel contains the voxel probabilities for one tract.

## 2.1. Preprocessing

While raw image values could have been used as input for our method, this would have restricted the method to the MRI acquisition used, not even allowing for slight variations in the acquisition setup without a complete retraining of the method. Moreover, for high angular resolution datasets, it would have resulted in an input image with a huge number of channels, creating high memory demand and slow file input/output during training. A more condensed representation of the data was chosen to mitigate this problem: TractSeg expects to receive the three principal fiber directions per voxel as input, thus requiring nine different input channels (three per principal direction). In this study, the principal directions were extracted using the multi-shell multi-tissue constrained spherical deconvolution (CSD) and peak extraction available in MRtrix (Jeurissen et al., 2014, Tournier et al., 2007) with a maximum number of three peaks per voxel. If a voxel contained only one fiber direction e.g. voxels in the corpus callosum, then the second and third peak are set to zero. The HCP images have a spatial resolution of 145x174x145 voxels. We cropped them to 144x144x144 without removing any brain tissue to make them fit to our network input size.

## 2.2. Convolutional Neural Network

The proposed 2D encoder-decoder FCNN architecture was inspired by the U-Net encoder-decoder architecture previously proposed by Ronneberger et al. (Ronneberger et al., 2015). Figure 2 details the proposed network setup.

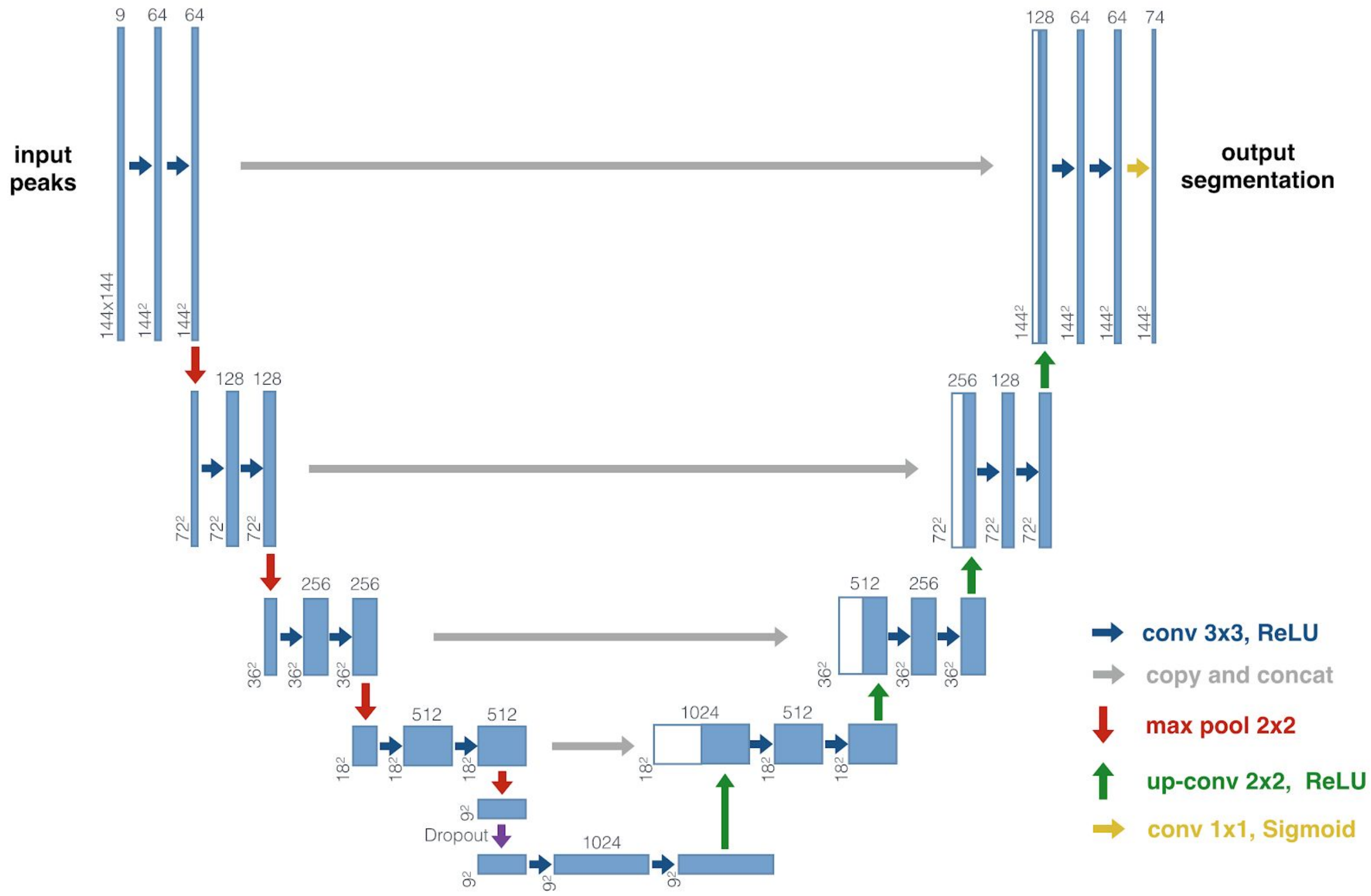

Figure 2: Proposed TractSeg FCNN architecture. Blue boxes represent multi-channel feature maps. White boxes show copied feature maps. The gray number on top of each box gives the number of channels, the x-y-size is given at lower left corner of each box. Network operations are represented by differently colored arrows.

The input for the proposed network is a 2D image at 144x144 voxels and 9 channels corresponding to the 3 peaks per voxel (each peak is represented by a 3D vector). The output is a multi-channel image (72 channels) with spatial dimensions of 144x144 voxels, where each channel contains the voxel probabilities for one tract. These probabilities were converted to binary segmentations by thresholding at 0.5. The approach enables for multi-label segmentations with several tracts sharing one voxel. This means that one network can learn all of the tracts simultaneously. To enable this, the softmax activation in the last layer was replaced by a sigmoid activation, which does not require the probabilities of all classes to sum to one. To avoid a downsized output in comparison to the input we padded with half the filter size (rounded down) on both sides. Given a filter size of 3 the padding was set to 1. This is also referred to as SAME padding (Dumoulin et al., 2016).

While in principle the U-Net architecture allows extensions to 3D image segmentation (Çiçek et al., 2016), we propose that 2D slices be used as input to boost memory efficiency and enable the data to be processed at the original HCP resolution. To still leverage the additional information provided by the third dimension, we randomly sampled 2D slices in three different orientations during training: axial, coronal and sagittal. This meant that our model learned to work with all three of these orientations. During inference three predictions per voxel per tract were generated, one for each orientation, resulting in an image with dimensions of 144x144x144x72x3 (after running our model 144*3 times) and, if we concatenate it along the last 2 dimensions, we get 144x144x144x216 (a 3D image with 216 channels). A second FCNN with three input channels per tract (i.e. 3*72=216 input channels) was trained to optimally merge the three predictions per voxel. However, as the second FCNN also works on 2D slices we can sample in three different orientations (axial, coronal, sagittal) again here. The second FCNN was thus also ran three times, once for every orientation, giving three predictions per voxel per tract. We simply took the mean to merge those to one final segmentation. The second FCNN is optional. We could have directly used the mean to merge the outputs from the first FCNN, but using the second FCNN allows a second network to learn which combination of slice orientations yields the best prediction for each tract or part of tract individually, thus further increasing segmentation quality.

## 2.3. Training

We trained our network using the binary cross-entropy loss. For a given target *t,* an output of the network *o* and *n* number of classes, the loss was calculated as follows:

$$loss(o,t) = -\frac{1}{n} \sum_{i=0}^{n} (t[i] * log(o[i]) + (1 - t[i]) * log(1 - o[i])) .$$

We used rectified linear units (ReLU) as nonlinearity (Nair and Hinton, 2010). Sigmoid activation functions were only used in the last layer. Sigmoid activation paired with binary cross-entropy enables multilabel segmentations. A learning rate of 0.002 was used and Adamax (Kingma and Ba, 2014) was implemented as an optimizer. The batch size was 56. We used dropout (Srivastava et al., 2014) with a probability of 0.4. All hyperparameters were optimized on a validation dataset independent of the final test dataset. The network weights of the epoch with the highest Dice score during validation were used for testing.

To improve the generalizability of TractSeg, we applied heavy data augmentation to the peak images during training[1]. The following transformations were applied to each training sample. The intensity of each transformation was varied randomly.

- Rotation by angle $\varphi_x \sim U[-\pi/4, \pi/4]$, $\varphi_y \sim U[-\pi/4, \pi/4]$, $\varphi_z \sim U[-\pi/4, \pi/4]$
- Elastic deformation with alpha and sigma $(\alpha, \sigma) \sim (U[90, 120], U[9, 11])$. A displacement vector is sampled for each voxel $d \sim U[-1, 1]$, which is then smoothed by a Gaussian filter with standard deviation $\sigma$ and finally scaled by $\alpha$.
- Displacement by $(\Delta x, \Delta y) \sim (U[-10, 10], U[-10, 10])$
- Zooming by a factor $\lambda \sim U[0.9, 1.5]$
- Resampling (to simulate lower image resolution) with factor $\lambda \sim U[0.5, 1]$
- Gaussian Noise with mean and variance $(\mu, \sigma) \sim (0, U[0.05])$
- Contrast augmentation $X_{augmented} = (X - avg(X)) * \beta + avg(X)$ with $\beta \sim U[0.7, 1.3]$
- Brightness augmentation $X_{augmented} = X * \gamma$ with $\gamma \sim U[0.7, 1.3]$

When training our network on peaks generated by the MRtrix multi-shell multi-tissue CSD method, we found that it did not work well on peaks generated by the standard MRtrix CSD method. In order to ensure our model worked well with all types of MRtrix peaks, we generated three peak images: (1) multi-shell multi-tissue CSD using all gradient directions, (2) standard CSD using only b=1000 s/mm$^2$ gradient directions, (3) standard CSD using only 12 gradient directions at b=1000 s/mm$^2$. During training, we randomly sampled from these three peak images, thus ensuring that our network worked well with all of them.

[1] https://github.com/MIC-DKFZ/batchgenerators

We trained for 500 epochs with each epoch corresponding to 162 batches. This means that over the course of the entire training, the network has seen 4,536,000 slices which have been randomly sampled from axial, coronal and sagittal orientations, randomly sampled from three different peak types and randomly permutated by the data augmentation transformations.

The results presented in section 3 were obtained using an implementation of the proposed method in Lasagne (Dieleman et al., 2015). Additionally we provide a Pytorch (pytorch.org) implementation.

## 2.4. Reference Segmentations

105 subjects from the Human Connectome Project (HCP) were used for the following experiments. The corresponding dMRI images have 1.25mm isotropic resolution and 270 gradient directions with 3 b-values (1000, 2000, 3000 s/mm$^2$) and 18 b=0 images (Sotiropoulos et al., 2013). We used the HCP data that had already been processed by the minimal preprocessing pipeline (e.g. distortion correction, motion correction, registration to MNI space and brain extraction had already been completed) (Glasser et al., 2013).
We semi-automatically generated binary reference segmentations for 72 major white matter tracts for each subject (figure 3). Those reference segmentations are used as labels for training and validating our network. We used a series of 5 processing steps (figure 4) to generate these reference segmentations:

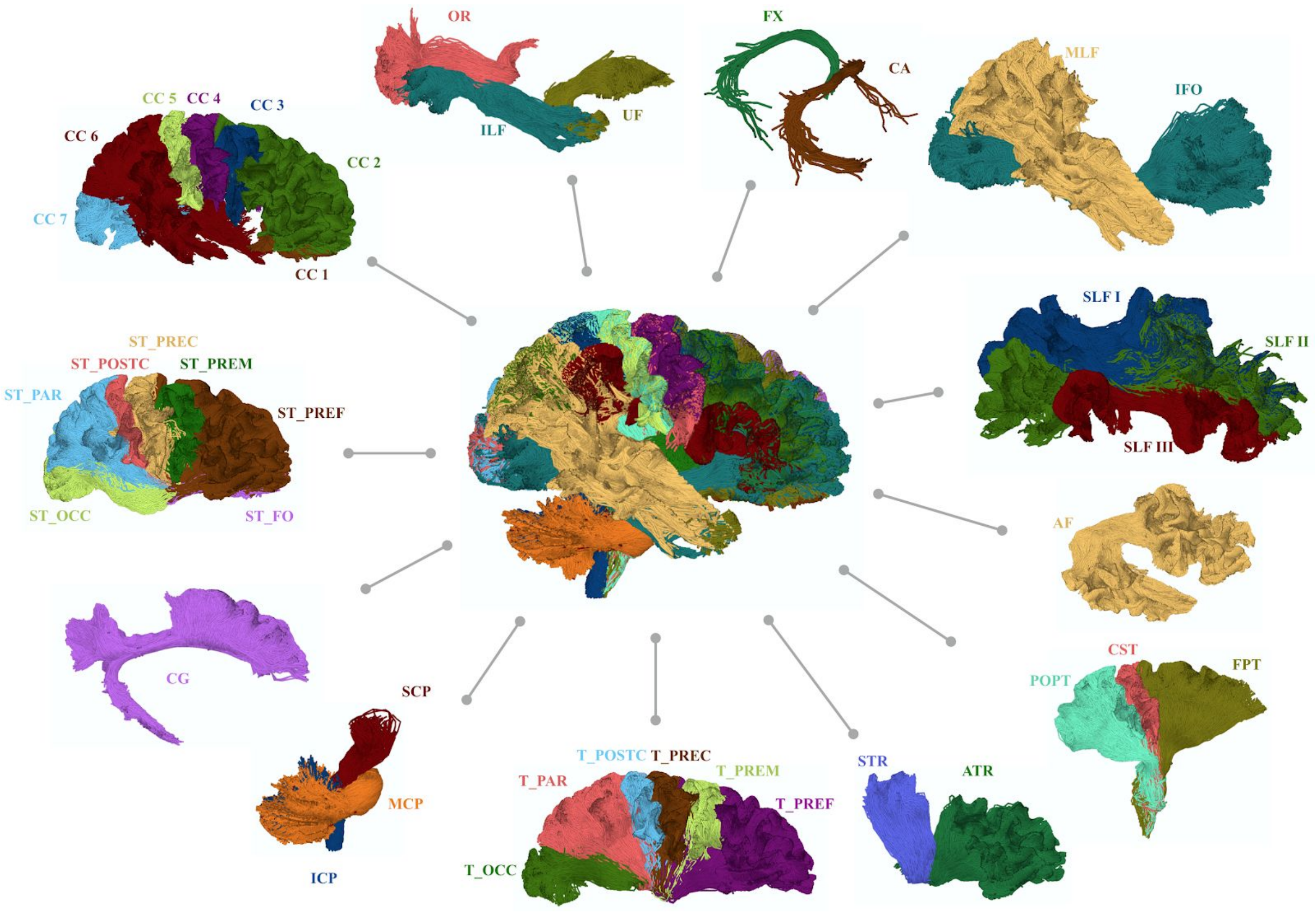

Figure 3: Overview of all 72 tracts. For tracts which exist in the left and the right hemisphere only the right one is shown. The following tracts are included: Arcuate fascicle (AF), Anterior thalamic radiation (ATR), Commissure anterior (CA),

Corpus callosum (Rostrum (CC 1), Genu (CC 2), Rostral body (CC 3), Anterior midbody (CC 4), Posterior midbody (CC 5), Isthmus (CC 6), Splenium (CC 7)), Cingulum (CG), Corticospinal tract (CST), Middle longitudinal fascicle (MLF), Fronto-pontine tract (FPT), Fornix (FX), Inferior cerebellar peduncle (ICP), Inferior occipito-frontal fascicle (IFO), Inferior longitudinal fascicle (ILF), Middle cerebellar peduncle (MCP), Optic radiation (OR), Parieto-occipital pontine (POPT), Superior cerebellar peduncle (SCP), Superior longitudinal fascicle I (SLF I), Superior longitudinal fascicle II (SLF II), Superior longitudinal fascicle III (SLF III), Superior thalamic radiation (STR), Uncinate fascicle (UF), Thalamo-prefrontal (T_PREF), Thalamo-premotor (T_PREM), Thalamo-precentral (T_PREC), Thalamo-postcentral (T_POSTC), Thalamo-parietal (T_PAR), Thalamo-occipital (T_OCC), Striato-fronto-orbital (ST_FO), Striato-prefrontal (ST_PREF), Striato-premotor (ST_PREM), Striato-precentral (ST_PREC), Striato-postcentral (ST_POSTC), Striato-parietal (ST_PAR), Striato-occipital (ST_OCC)

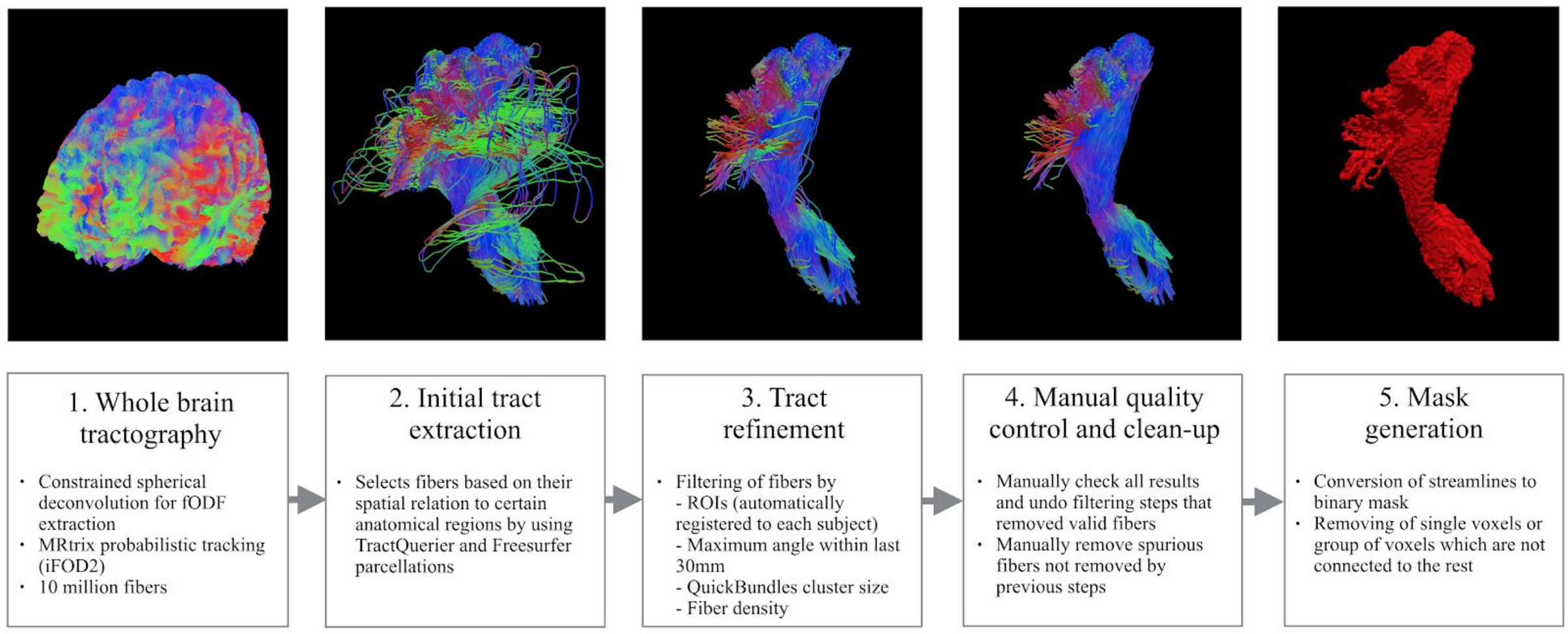

Figure 4: Examplary depiction of the semi-automatic dissection pipeline of a total of 72 reference tracts in a cohort of 105 subjects. Images show the left CST of subject 898176.

**Step 1 - Tractography.** Multi-shell multi-tissue constrained spherical deconvolution was used to extract the fODF and anatomically constrained probabilistic tractography (iFOD2) for the whole brain was performed using MRtrix (Tournier et al., 2010) to generate 10 million fibers with a minimum length of 40mm. By generating such a large number of fibers, we were able to reconstruct difficult tracts like the anterior commissure (CA) or the lateral projections of the corticospinal tract (CST) with high completeness in a majority of the subjects. Extracting the fibers that run completely from the left cortex through the CA to the right cortex is difficult if a smaller number of fibers is used. The high fiber count, however, comes at the cost of many false positives which we eliminated using the following steps. Seeds were randomly placed within the brain mask coming with the HCP data. Streamlines were cropped at the gray-matter-white-matter interface.
Anatomically constrained tractography yielded adequate results for the majority of tracts. However, the anatomical constraints did not work well for tracts containing very thin passages like the CA, uncinate fascicle (UF) or inferior occipito-frontal fascicle (IFO). They are based on a gray matter segmentation which is based on the T1 image and as a result they do not necessarily fit perfectly to the diffusion weighted image. If a tract is only a few voxels wide and the gray matter segmentation is not perfect, some critical parts of the tract will be cut off because they are counted as gray matter. Therefore we used tracking without anatomical constraints for the CA, UF and IFO.

**Step 2 - Initial tract extraction.** We applied TractQuerier (Wassermann et al., 2016) to extract a first approximation of each tract and to make sure streamlines end in the correct cortex regions. Our queries are based on the queries provided by Wassermann et al. (2016) with minor adaptations for better results with our dataset. The queries used in this work are available online[2]. As was the case in Wassermann et al. (2016), the FreeSurfer cortical parcellations were used.

**Step 3 - Tract refinement.** The results of TractQuerier contained a considerable amount of false positives (see figure 4). This meant that further filtering was needed. We manually defined exclusion and inclusion ROIs to exclude false positives (Stieltjes et al., 2013). These ROIs s were transferred between subjects using diffeomorphic registration of the fractional anisotropy (FA) maps.
We also removed streamlines that run back and forth in the target volume, making 180 degree turns within a length of less than 30 mm. More spurious fibers were filtered by running QuickBundles clustering (Garyfallidis et al., 2012) and removing clusters that only contained a small number of fibers, according to thresholds which varied for each tract. This approach was used very cautiously as it also tends to remove valid structures such as the lateral projections of the CST, which are rather sparse in streamline space and therefore typically consist of small clusters. Finally, we removed all streamlines that run through voxels with low streamline density (threshold differs for each tract).

**Step 4 - Manual quality control, clean-up.** The described steps for automated tract refinement occasionally removed some of the valid parts of a tract. We manually inspected the results of all filtering steps in all subjects and skipped some of the filters if they removed too many valid fibers. After this process was complete, there were streamlines still remaining which could only be filtered by drawing individual exclusion ROIs manually (Stieltjes et al., 2013). The UF, for example, tended to contain parts of the CA. As a result of inaccurate Freesurfer parcellations, those could not be properly excluded in all subjects with TractQuerier. The necessary exclusion ROI had to be drawn around the valid part of the UF with minimal margins. Since this required a highly accurate placement, the ROI had to be drawn manually for each subject instead of using the same ROI for all subjects.

**Step 5 - Mask Generation.** This semi-automatic process resulted in high quality dissections of a total of 72 tracts in in a cohort of 105 subjects. From the final sets of streamlines, we created binary tract masks. These binary masks were analyzed for connected regions and only the biggest connected regions were retained, thus filtering out single voxels and small groups of voxels that were not connected to the rest.

These binary masks now serve as reference segmentations and are used for training and testing of the proposed segmentation approach.

## 2.5. Reference methods

The following automatic tract segmentation methods were used as a benchmark. All these methods are openly available and do not need any proprietary software to run. The methods include clustering-based as well as ROI-based approaches. We give an outline of how they work (1.) and how we applied them (2.) here.

**RecoBundles**: 1. Given streamlines of a reference tract in a reference subject, RecoBundles (Garyfallidis et al., 2017) can be used to find the corresponding streamlines in a new subject. 2. We randomly picked 5 reference subjects from the HCP data (each containing the 72 reference tracts we had previously extracted (see 2.4)). Due to the long runtime for RecoBundles, a higher number of reference subjects was not feasible. Then we ran RecoBundles 5 times for the new subject (once for each reference subject) using the default RecoBundles parameters. This resulted in 5 extractions of each tract in the new subject. To get a final segmentation, we took the mean of those 5 extractions.

[2] https://github.com/MIC-DKFZ/TractSeg/blob/master/examples/resources/WMQL_Queries.qry

**TRACULA**: 1. TRACULA (Yendiki et al., 2011) uses probabilistic tracking and an atlas of the underlying anatomy to segment tracts. 2.Running TRACULA for each subject produced a probability map for each tract. We selected all voxels with a probability of greater than 0 to create a binary segmentation. The original paper created a binary segmentation by masking out all values below 20% of the maximum. However, this leaves only the very core of every tract, thus creating many false negatives. Segmenting all values with probability greater than 0 gives better results compared to our reference tracts. TRACULA does not support all of the 72 tracts that we use; it only supports 18. Out of the 18, only 8 tracts match our 72 tracts exactly. Quantitative comparison was therefore only performed on these 8 tracts. A GPU implementation is available for *bedpost,*which is part of the TRACULA pipeline which we used (Hernandez et al., 2013).

**WhiteMatterAnalysis (WMA)**: 1. WMA (O'Donnell et al., 2012; O'Donnell and Westin, 2007) clusters streamlines across subjects and generates a cluster atlas out of these. Clusters in the atlas are then assigned to certain anatomical tracts. Automated tract delineation is made possible by registering new subjects to this atlas. 2. WMA comes with a pre-trained cluster atlas (800 clusters) and predefined mapping of clusters to anatomical tracts (O'Donnell et al., 2016). However, this mapping only includes 10 of the 72 tracts we use. We manually optimized these predefined mappings to better align them with our reference tracts and added the CA to the mapping. During this process, we were inherently limited by the finite set of distinct clusters offered by the atlas. We chose not to extend the mapping to the remaining 61 tracts as it would have required considerable manual effort, given the amount of clusters and tracts. Evaluation was therefore only carried out on these 11 tracts. WMA is very memory intensive. Clustering a tractogram with 1 million fibers requires around 60GB of memory. Therefore we were not able to use our 10 million fiber tractogram directly, but instead used a subset with 500,000 fibers. This reduced the memory consumption to around 30GB. The outlier removal provided by WMA was used to remove spurious fibers.

**TractQuerier**: 1. TractQuerier (Wassermann et al., 2016) extracts tracts based on the regions the streamlines have to start at, end at and (not) run through. 2. We compared our method to the output from TractQuerier (step 2 from the reference tract extraction pipeline) without any further post-processing.

In addition to these openly available reference methods, we implemented two additional reference methods in order to provide further insights into the performance of TractSeg:

**Atlas registration (Atlas)**: 1. Several subjects can be averaged to an atlas which can then be registered to new subjects to segment structures. 2. We split our reference data (see 2.4) into training and testing data, using the same 5-fold cross-validation as used for the evaluation of our proposed method. The training data was used to create a tract atlas. Firstly, we registered all subjects to a random subject using symmetric diffeomorphic registration implemented in DIPY (Avants et al., 2009, Garyfallidis et al., 2014). Registration was performed based on the FA maps of each image.

After registration, the FA maps of all images were averaged. Then, in a second iteration all images were registered to this mean FA image. This two-stage approach limits the bias introduced by the initial subject choice in the first iteration. The tract atlas thus contained the tract masks for all 72 reference tracts (see 2.4). For each tract, we took the mean over all subjects, which produced a probability map. We thresholded the probability map at 0.5 to create a final binary atlas. During test time, the atlas was registered to the subjects of interest, yielding a binary mask for each tract in subject space.

**Multiple mask registration (Multi-Mask)**: 1. Using an atlas can blur some of the details as it is based on group averages. The blurring can be reduced to some extent by registering the masks of single training subjects to a test subject instead of an averaged atlas. 2. The same 5 reference subjects as those selected for RecoBundles were used. To segment the tracts in a new subject, we registered each of the 5 reference subjects to the new subject (symmetric diffeomorphic registration of the FA maps) and averaged the tract masks (from the reference tracts) of all 5 reference subjects. Finally, we thresholded this average at 0.5 to produce a binary mask for each tract in the space of the new subject. This differs from the *Atlas registration* method in that the reference subjects are directly registered to subject space and are merged (1 registration) instead of first being registered to atlas space, then merging and being registered to subject space (2 registrations needed). Moreover, *Atlas Registration* uses 63 subjects while *Multi-Mask* only uses 5.

# 3. Experiments and results

The high quality HCP data (*HCP Quality*) was used in order to extract the best possible reference tracts. However, in clinical routine, faster MRI protocols are used which result in lower quality data. To test how TractSeg performs on clinical quality data, we downsampled the HCP data to 2.5mm isotropic resolution and removed all but 32 weighted volumes at b=1000 s/mm$^2$. The reference tracts from the *HCP Quality* dataset were reused as our reference tracts here. This provides high quality reference tracts for the low quality data, thus allowing proper evaluation. We call this dataset *Clinical Quality*. As this dataset only has one b-value shell, we can not use multi-shell CSD as we did for the *HCP Quality* data. Instead MRtrix standard CSD was employed to generate the peaks of the fODF. 5 fold cross validation was used, i.e. 63 training subjects, 21 validation subjects (best epoch selection) and 21 test subjects per fold.

We used the Dice score (Taha et al., 2015) as our evaluation metric, which is one of the most popular measures for evaluating segmentation performance. We calculated the Dice for each subject between each of the 72 reference tracts and the respective prediction of either our proposed method or one of the reference methods (e.g. RecoBundles). Then we averaged the Dice results for all 72 tracts to get one final Dice score per subject per method.

The Wilcoxon signed-rank test (Wilcoxon, 1945) was used to test for statistical significance. For multiple testing, we applied the Bonferroni correction. Moreover, we evaluated the runtime of the different methods, the influence of different orientation fusion strategies and how TractSeg generalizes to non-HCP data.

## 3.1. Quantitative evaluation

On the *HCP Quality* data, looking at the mean Dice score over all 72 reference tracts, TractSeg significantly (p<0.001) outperformed the second best method by 9 Dice points and on average performed 19 Dice points better than the other reference methods (figure 5a). On the *Clinical Quality* data, TractSeg significantly (p<0.001) outperformed the second best method by 14 Dice points and performed 22 Dice points better than the reference methods on average (figure 5b). In general, TractSeg was less affected by the quality loss in the clinical data than the reference methods. Please note that *WMA* and *TRACULA* only provide segmentations for a subset of the 72 tracts. The reported scores for *WMA* and *TRACULA* are restricted to this subset.

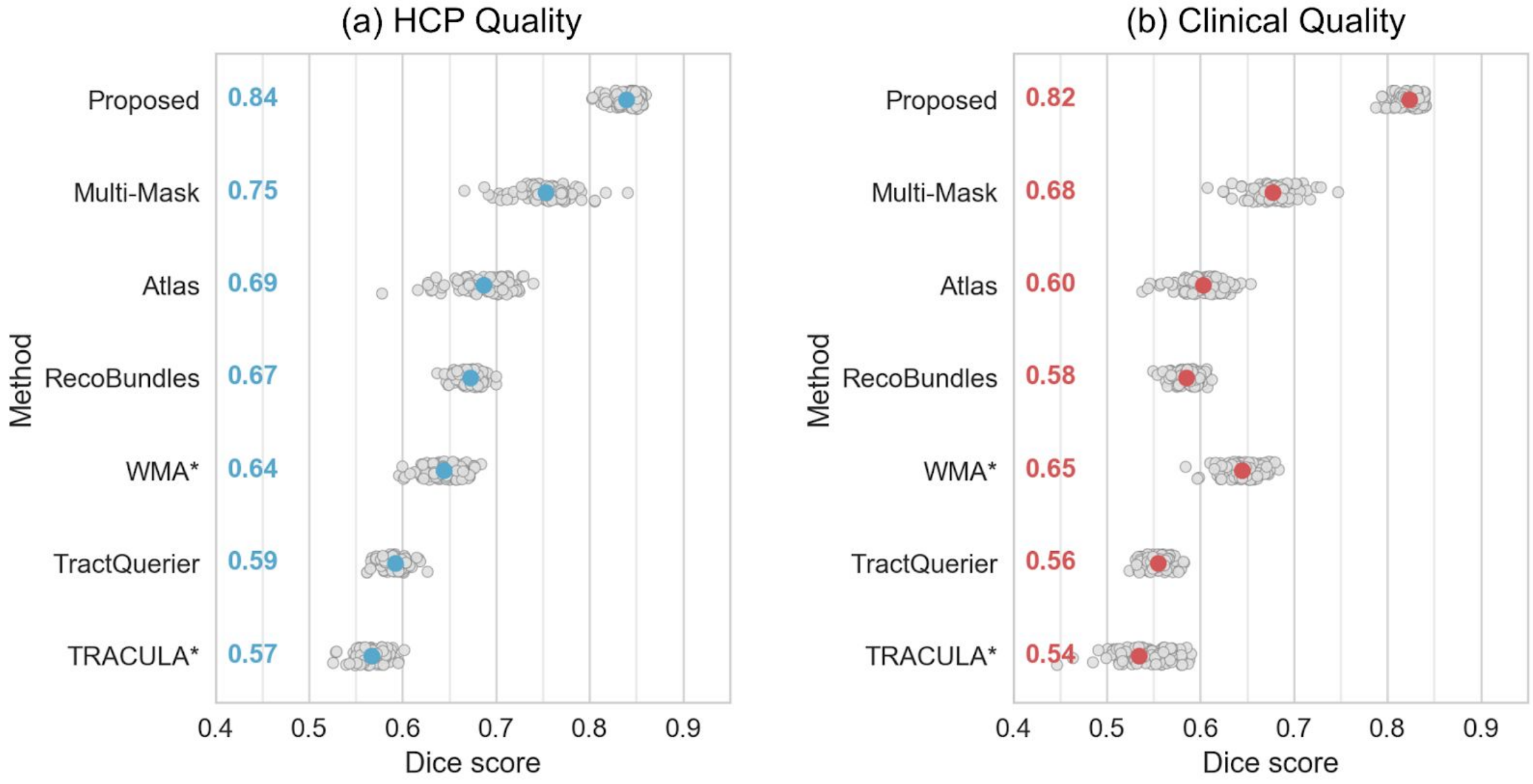

Figure 5: Results on (a) the *HCP Quality* dataset and (b) the *Clinical Quality* dataset with a gray dot per subject (mean over all tracts) and a colored dot for the mean over all subjects. *Proposed*: Our method; *Multi-Mask*: Multiple mask registration; *Atlas*: Atlas registration; *WMA*: WhiteMatterAnalysis.
*: *WMA* and *TRACULA* do not provide segmentations for all tracts, only for a subset. The score is the mean over the tracts in the subset

Figure 6 shows the segmentation performance for all methods for each individual tract. TractSeg showed quite consistent performance. All tracts had a Dice score over 0.75, except for the fornix (FX) and CA. For these two tracts, the Dice score was noticeably reduced, which is probably caused by the thin and therefore hard to reconstruct shape of these tracts. Moreover, due to limited sensitivity of the initial tractograms, the CA and FX were not perfectly reconstructed in the reference tracts for some subjects but our model managed to generate a segmentation which was better than the reference segmentation (see section 3.2). Therefore good segmentations also received bad Dice scores if the reference segmentation was incomplete. The reference methods showed a similar accuracy gap between the FX and CA and the remaining tracts. Overall, TractSeg clearly outperformed the reference methods.

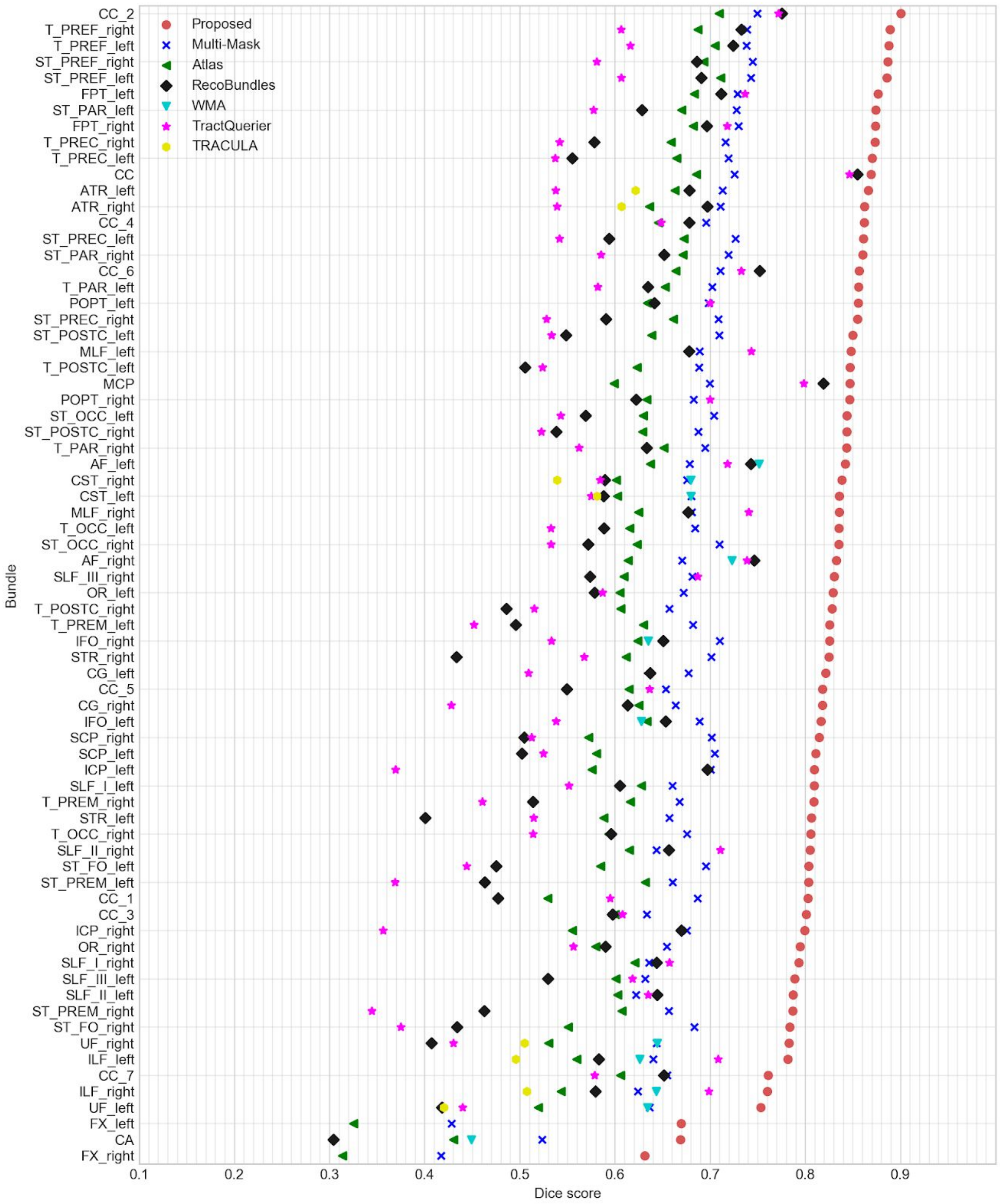

Figure 6: Dice scores for all 72 tracts on the *Clinical Quality* dataset for our proposed method and all reference methods sorted by score. The full name of each tract can be seen in figure 3.

## 3.2. Qualitative evaluation

For the qualitative evaluation, one subject (623844) was selected from the test set. We chose a subject whose Dice scores were closest to the mean Dice scores for the entire datasets to make the subject representative for the entire dataset. Since the scope of this manuscript does not allow us to show results for all 72 tracts, we selected three tracts that represent the different degrees of reconstruction difficulty (Maier-Hein et al., 2017): the IFO, CST and CA (results for all tracts can be accessed online [3]). The IFO is a tract that is fairly easy to reconstruct, which is reflected by its consistently good scores for all methods. The CST is more difficult to reconstruct. Its beginning at the brain stem is easy to reconstruct but as the fibers get closer to the cortex, they start to fan out. Finding these lateral projections is more difficult. Finally, the CA is tract that is difficult to reconstruct. Due to its very thin body, it is hard to find fibers that run the entire way from the right to the left temporal lobe. The CA is one of the tracts with the lowest performance out of all of the methods.

As can be seen in Figure 7, the *Proposed* method yielded adequate reconstructions on all three tracts. *Multi-Mask* registration also produced convincing results on all tracts. Atlas registration results looked convincing on the 3D image but included segments passing gray matter and also exiting the brain mask, which is obviously wrong. *RecoBundles* oversegmented the CST to neighbouring gyri and did not manage to reconstruct the CA completely but showed good results for the IFO. *TractQuerier* did not properly segment any of the example tracts. As it defines tracts mainly by their endpoints, it leaves much room for wrong turns between the start and end points. *TractQuerier* extracts a lot of false positives, especially when using probabilistic tracking. The CA cannot be properly reconstructed with TractQuerier as the default Freesurfer parcellation is not precise enough for the small parts of the CA. *TRACULA* yielded a good reconstruction of the stem of the CST, but missed the lateral projections. CA and IFO are not supported by *TRACULA*. White Matter Analysis (WMA) kept a lot of false positives in the CST and CA but showed good results for the IFO.

In a similar comparison of results on the *Clinical Quality* dataset (figure 8), as expected, all approaches yielded segmentations of lower quality compared to the respective segmentation on the *HCP Quality* data, to varying extents, as can be seen in figure 8. The registration-based approaches also tended to segment gray matter or regions outside of the brain while the streamline based approaches suffered from an increased number of false positives. TractSeg seems to be less affected by the lower image quality than most of the benchmark methods, as shown by the quantitative analysis.

In general, it can be observed that tracts tend to look good on 3D image but show significant shortcomings on a detailed 2D image. Unfortunately, many of the relevant publications only show qualitative results using single 3D images. The segmentations of all methods and all tracts of this subject are available online[3].

[3] https://doi.org/10.5281/zenodo.1154877

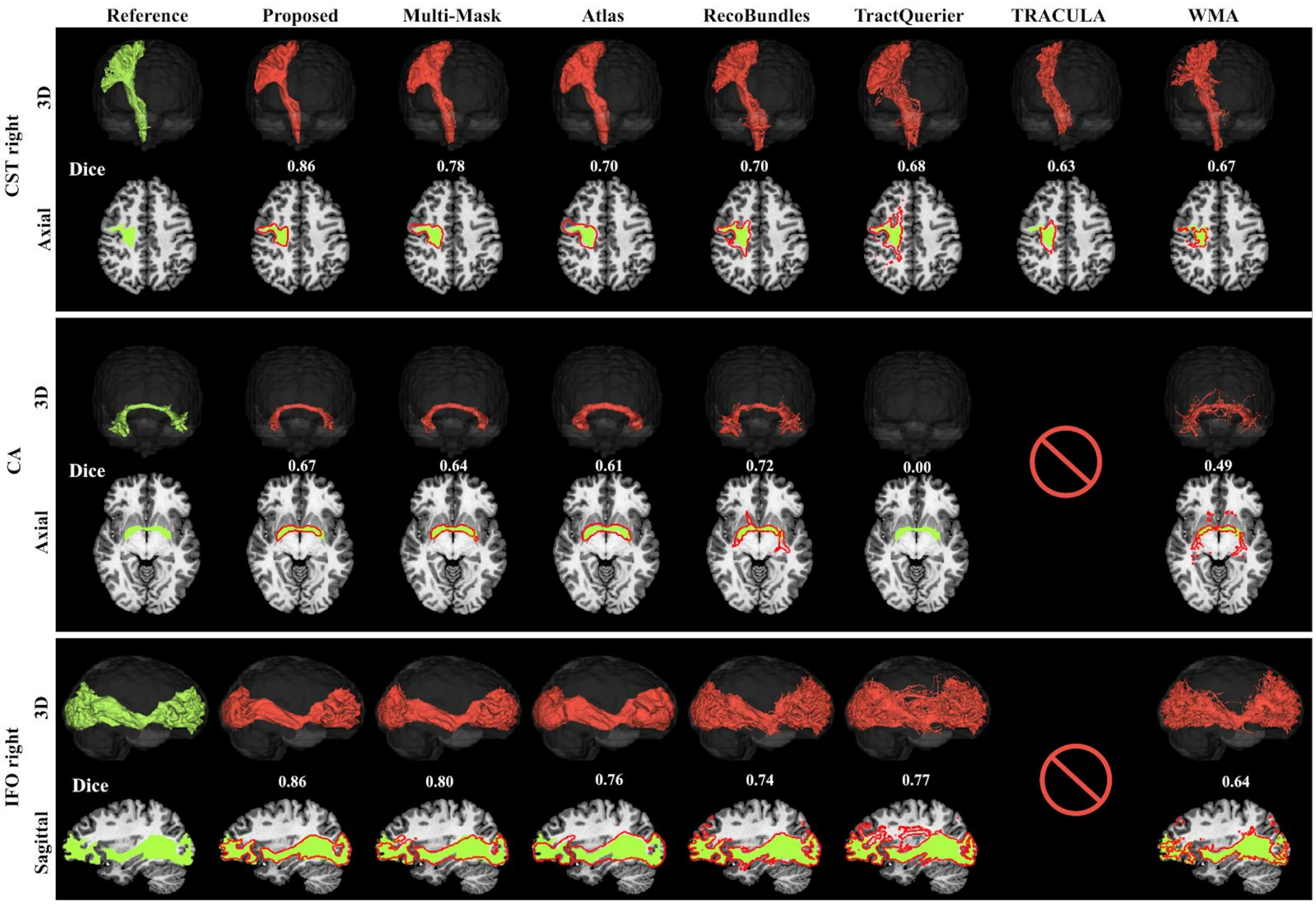

Figure 7: Qualitative comparison of results on *HCP Quality* test set: reconstruction of right corticospinal tract (CST), commissure anterior (CA) and right inferior occipito-frontal fascicle (IFO) on one random subject. *Green* shows the reference tract and *Red* shows the segmentation of the respective method.

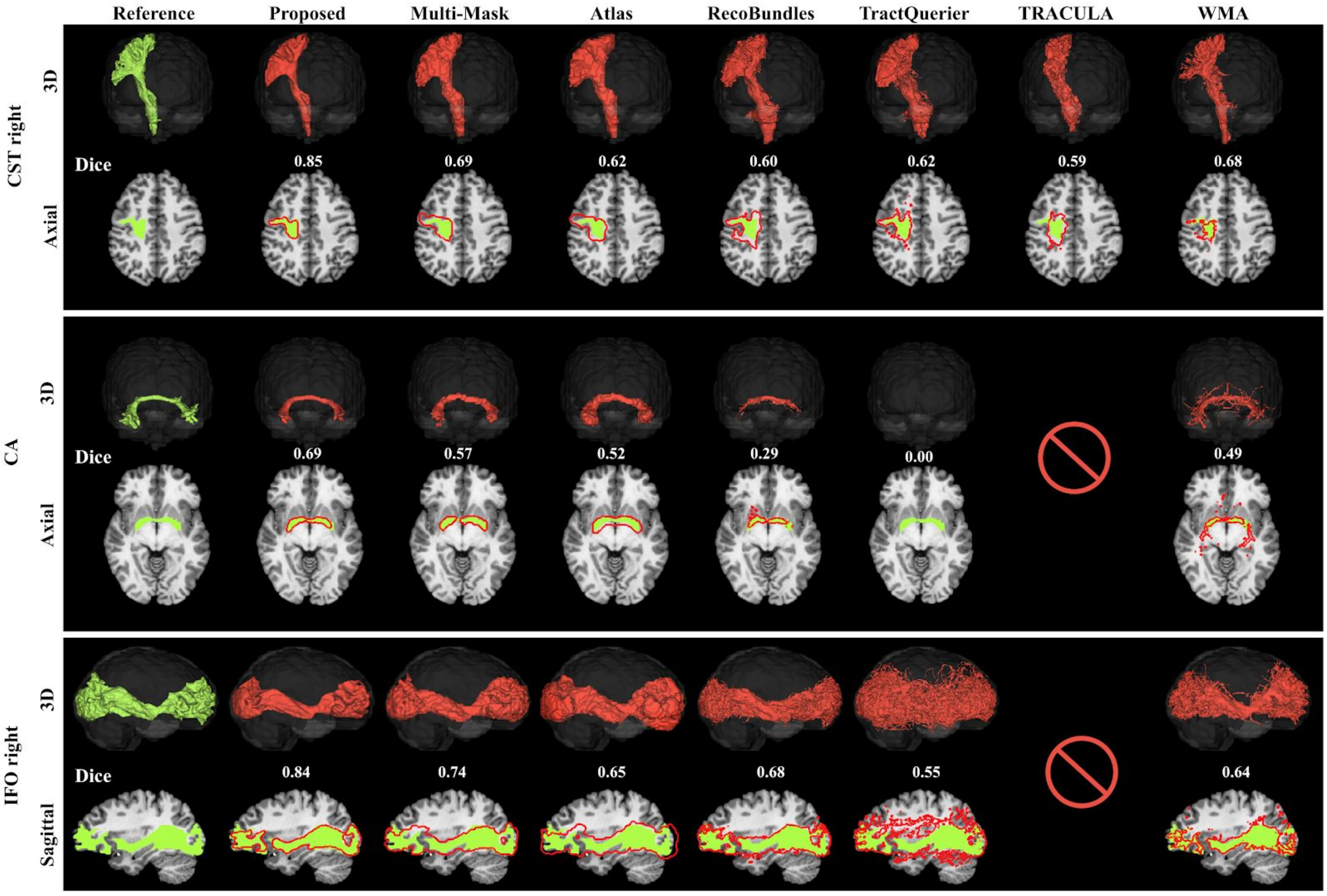

Figure 8: Qualitative comparison of results on *Clinical Quality* test set: reconstruction of right corticospinal tract (CST), commissure anterior (CA) and right inferior occipito-frontal fascicle (IFO) on one random subject. *Green* shows the reference tract and *Red* shows the segmentation of the respective method.

In some of the reference subjects, some of the more difficult tracts, such as the CA and the lateral projections of the CST, were incompletely reconstructed. The original tractogram contained a few sparse fibers for those regions, indicating the presence of the tract, however, when applying further constraints (ending in the correct cortex regions, angular threshold, fiber density) those sparse fibers were completely removed. Despite this noise in the training data, the proposed method managed to learn a complete representation of each tract. This enabled our model to completely segment these tracts, even on subjects in whom the reference tracts are incomplete (see figure 9).

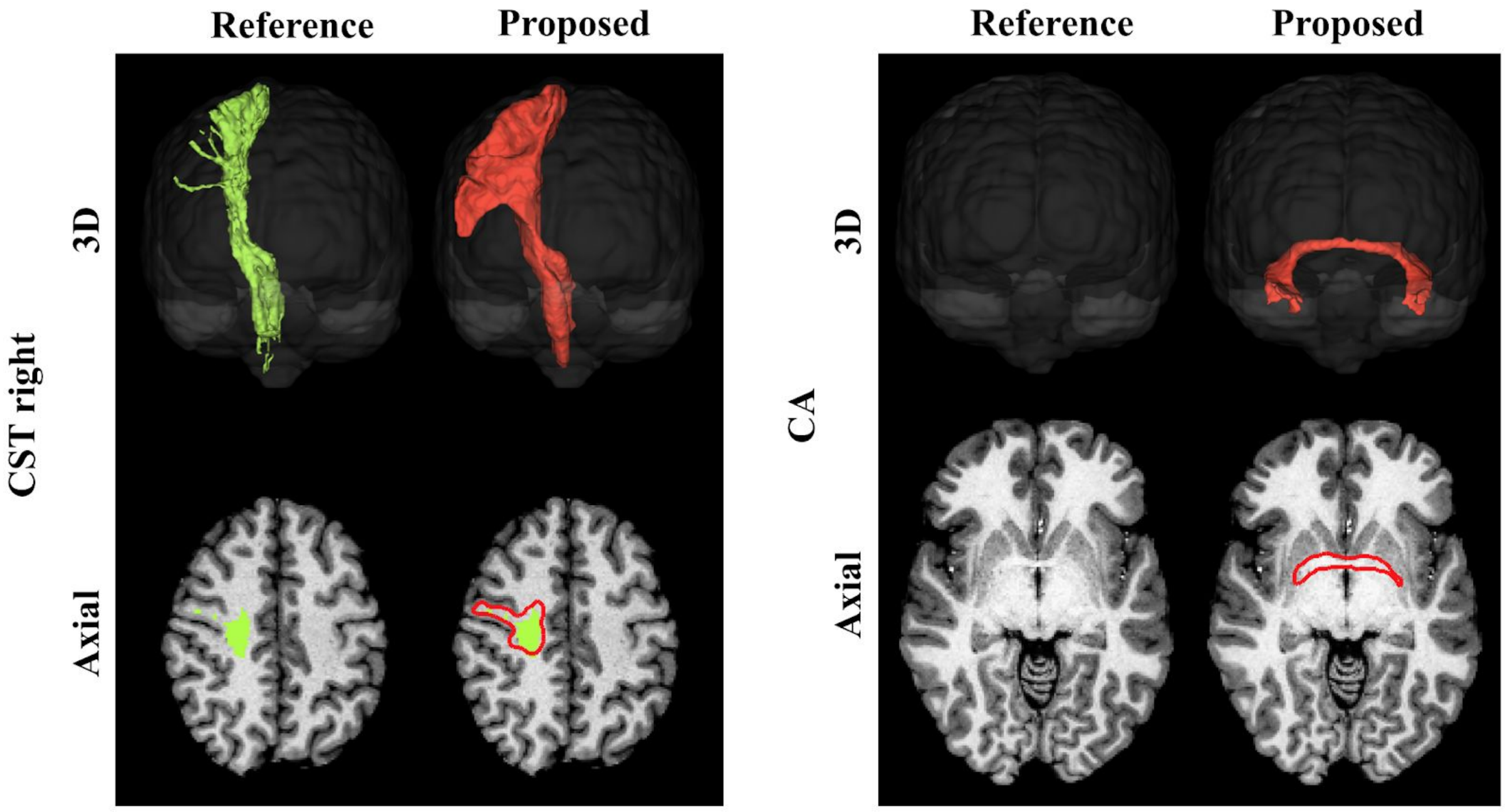

Figure 9: The proposed method creates anatomically reasonable segmentations even on subjects where the reference segmentations are incomplete due to limitations in the extraction pipeline. The presented examples are from the *HCP Quality* dataset (subjects 979984 and 959574).

## 3.3. Transferability between scanners and acquisition settings

To test the capability of TractSeg to generalize beyond HCP, which it was trained on, we applied TractSeg to nine differently acquired datasets (see table 1). These nine datasets represent a wide variety of data: Different scanners, different spatial resolutions, different b-values, different number of gradients, healthy and diseased, normal und abnormal brain anatomy.

Table 1: Acquisition parameters of additional non-HCP test datasets.

| **Project** | **Pathology** | **Resolution (isotropic)** | **b-Values** | **Preprocessing** | **Scanner field strength** |
|---|---|---|---|---|---|
| TRACED[4] | healthy | 2.5mm | 3x b=0mm/s$^2$<br>20x b=1000mm/s$^2$<br>48x b=2000mm/s$^2$ | denoising,<br>eddy current and motion correction | 3T |

[4] https://my.vanderbilt.edu/ismrmtraced2017/

| | | | | | |
|---|---|---|---|---|---|
| | | | 64x b=3000mm/s$^2$ | | |
| Internal (Healthy) | healthy | 2.5mm | 1x b=0mm/s$^2$<br>81x b=1000mm/s$^2$<br>81x b=2000mm/s$^2$<br>81x b=3000mm/s$^2$ | denoising,<br>eddy current and motion correction | 3T |
| BrainGluSchi (Bustillo et al., 2016) | healthy | 2.0mm | 5x b=0mm/s$^2$<br>30x b=800mm/s$^2$ | denoising,<br>eddy current and motion correction | 3T |
| Stanford_hardi (Rokem et al., 2013) | healthy | 2.0mm | 10x b=0mm/s$^2$<br>150x b=2000mm/s$^2$ | denoising,<br>eddy current and motion correction | 3T |
| Sherbrooke_3shell[5] | healthy | 2.5mm | 1x b=0mm/s$^2$<br>64x b=1000mm/s$^2$<br>64x b=2000mm/s$^2$<br>64x b=3500mm/s$^2$ | denoising,<br>eddy current and motion correction | 3T |
| COBRE (Çetin et al., 2014) | schizophrenia, enlarged ventricles | 2.0mm | 5x b=0mm/s$^2$<br>30x b=800mm/s$^2$ | denoising,<br>eddy current and motion correction | 3T |
| SoftSigns (Hirjak et al., 2017) | neurological soft signs | 2.5mm | 1x b=0mm/s$^2$<br>81x b=1000mm/s$^2$ | denoising,<br>eddy current and motion correction | 3T |
| Internal (Autism) | autism spectrum disorder | 2.5mm | 5x b=0mm/s$^2$<br>60x b=1000mm/s$^2$ | denoising,<br>eddy current and motion correction | 3T |
| Internal (Schizophrenia) | schizophrenia | 1.7mm | 3x b=0mm/s$^2$<br>60x b=1500mm/s$^2$ | denoising,<br>eddy current and motion correction | 3T |

An expert manually dissected the three tracts already shown in the above qualitative evaluation (CST, CA and IFO) from one randomly chosen subject from each dataset shown in Table 1 (from the COBRE dataset we choose 3 subjects: one healthy, one with schizophrenia and one with schizophrenia and enlarged ventricles). Visual comparisons were performed between manual dissections and TractSeg as well as the previously introduced reference methods *Multi-Mask, Atlas, RecoBundles* and *TractQuerier*. Methods depending on reference data (*TractSeg*, *Multi-Mask*, *Atlas*, *RecoBundles*) were provided with our HCP reference data. All subjects were rigidly registered to MNI space. This is not required for TractSeg to work. TractSeg only requires that the left/right, front/back and up/down orientation of the images are the same as for the HCP data (i.e. images are not *mirrored*). Rigid registration to MNI space is an easy way to ensure this.

TractSeg showed anatomically plausible results for all subjects and most of the tracts. Only the CA and the Fornix (FX) were not completely reconstructed in around half of the subjects. We observed partly incomplete manual reference dissections in these areas as well, indicating that the size of these very thin structures is reaching the resolution limit of the underlying imaging acquisition. More complete results could also be achieved by lowering the threshold for converting the CA and FX probability maps, which are generated by TractSeg, to binary maps. This would increase sensitivity at the cost of specificity. Figures 10 and 11 show exemplary results for two of the subjects. The other nine subjects can be found in the supplementary materials.

[5] http://nipy.org/dipy/reference/dipy.data.html#fetch-sherbrooke-3shell

Figure 10 shows the results for one healthy subject from the COBRE dataset (2mm isotropic resolution, 30x b=800mm/s$^2$). For the CST and IFO TractSeg yielded segmentations similar to the manual dissection. For the CA TractSeg missed parts but also the manual dissection was not able to reconstruct the CA completely. *Multi-Mask, Atlas* and *TractQuerier* showed more complete results however this comes at the price of severe over-segmentation (false positives). For the CST and IFO, Multi-Mask and Atlas even segmented non-brain areas and for the CA they segment major areas around the CA. The CA was not reconstructed at all by *TractQuerier*. *RecoBundles* also over-segmented the CST, but missed parts of the CA and IFO.

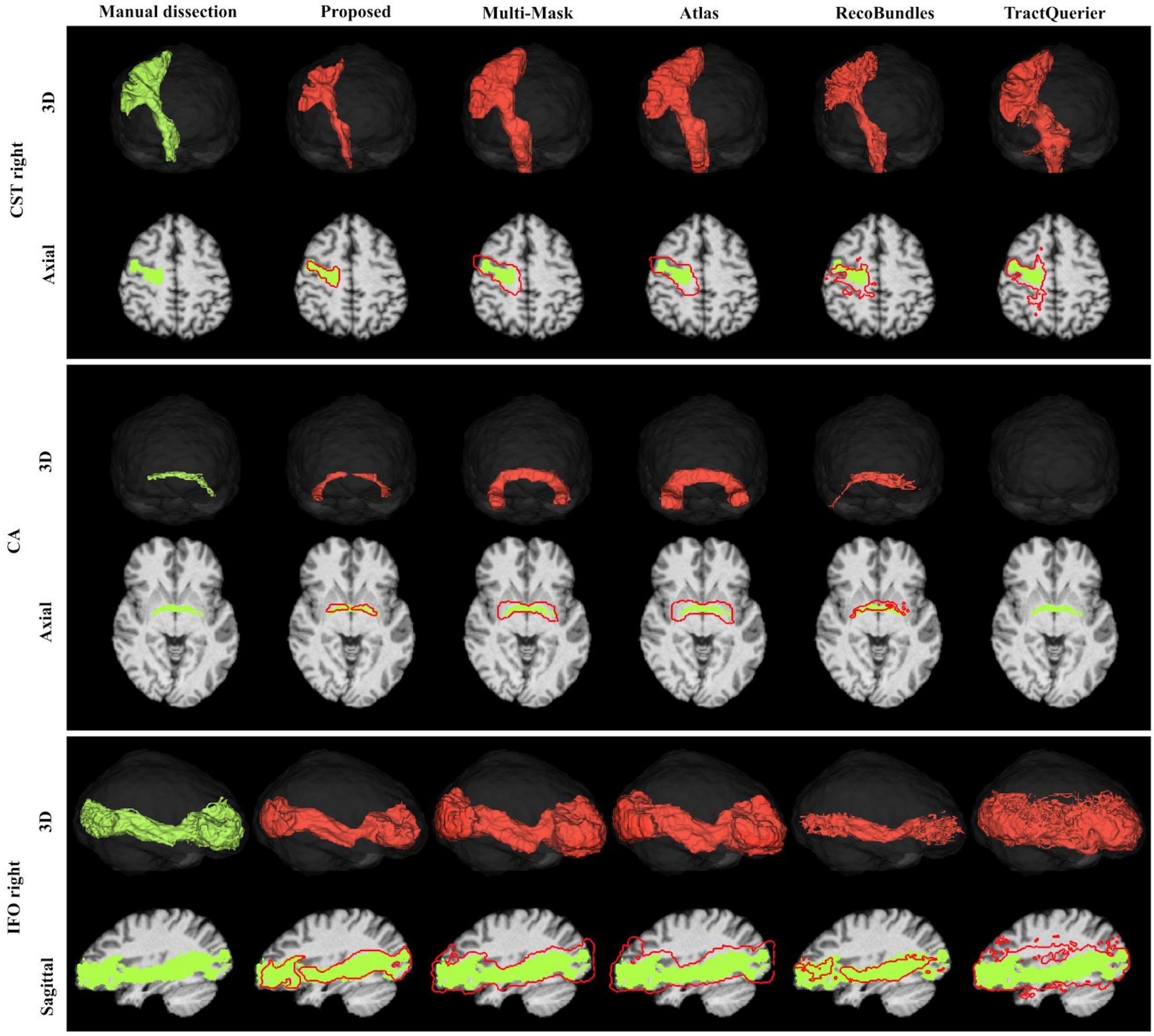

Figure 10: Qualitative comparison of results on one healthy subject from the COBRE dataset while being trained on the HCP dataset: reconstruction of right corticospinal tract (CST), commissure anterior (CA) and right inferior occipito-frontal fascicle (IFO).

Figure 11 shows the results for a schizophrenia patient with abnormally large ventricles from the COBRE dataset. Even though TractSeg has only seen healthy subjects with normally sized ventricles during training it managed to properly segment the main pathway of the CST. It missed small parts of the lateral projections of the CST but the same is true for the manual dissection. *Atlas* shows complete results but again at the cost of heavy over-segmentation even of parts of the ventricles.

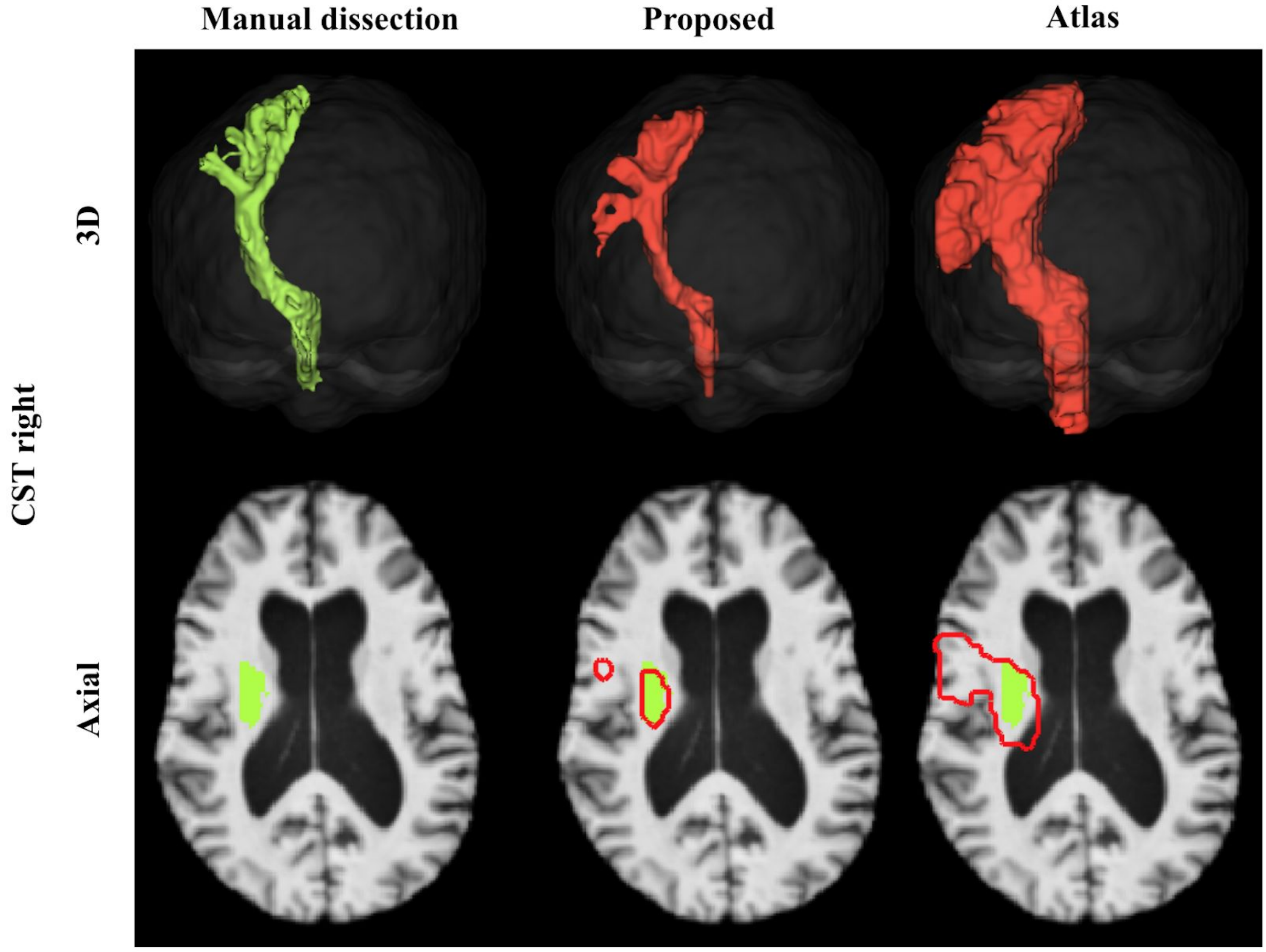

Figure 11: Qualitative comparison of results on one subject with schizophrenia and enlarged ventricles: reconstruction of right corticospinal tract (CST).

## 3.3. Fusion strategies

As described in section 2, we ran our model once for each orientation (axial, sagittal and coronal). This results in 3 predictions for each test image, which was merged into one final segmentation using a second FCNN. However, using two FCNNs instead of just one requires more resources during training and increases the runtime for inference. Therefore, also two simpler fusion strategies were also evaluated: Taking the *Mean* and *Majority Voting* instead of running the *Second FCNN*.

Figure 12 shows the results: *Mean* worked slightly better ($p<0.001$) than *Majority Voting*. On *HCP Quality, Mean* even worked slightly better ($p<0.001$) than the *Second FCNN* approach (0.2 Dice points). However, on *Clinical Quality, Mean* performed significantly worse ($p<0.001$) than the *Second FCNN* (1.9 Dice points). Thus, in practical applications, the benefits of improved training and inference performance of the *Mean* fusion strategy may outweigh the gains in performance achieved by the *Second FCNN*.

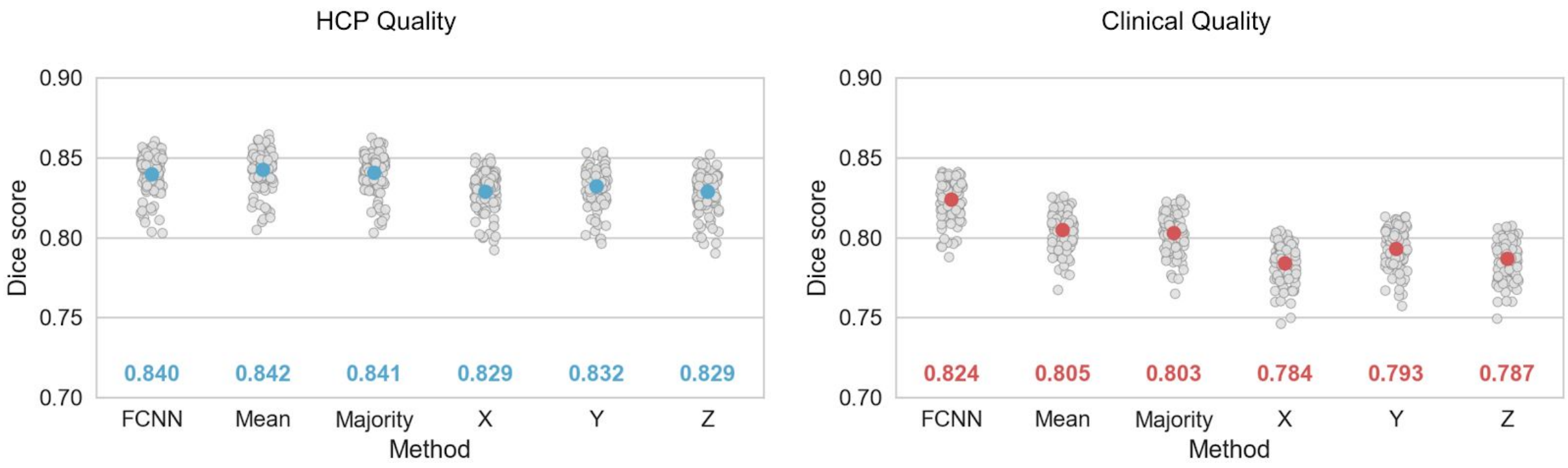

Figure 12: Results of different fusion strategies: *FCNN*: Second FCNN that received predictions from 3 slice orientations as input and produced the final output; *Mean*: The mean taken; *Majority*: Majority voting; *X,Y,Z*: No fusion, only the respective slice orientation.
Results show the mean Dice score over all tracts. Each subject is represented by one gray dot. The mean over all subjects is shown by the colored dot and the colored number.

Figure 13 shows the results of the different fusion strategies for all 72 tracts on the *Clinical Quality* dataset. *Second FCNN* consistently outperforms the simpler *Mean* fusion strategy, especially on the FX. *Y* constantly performs better than *X* or *Z* for the individual orientations.

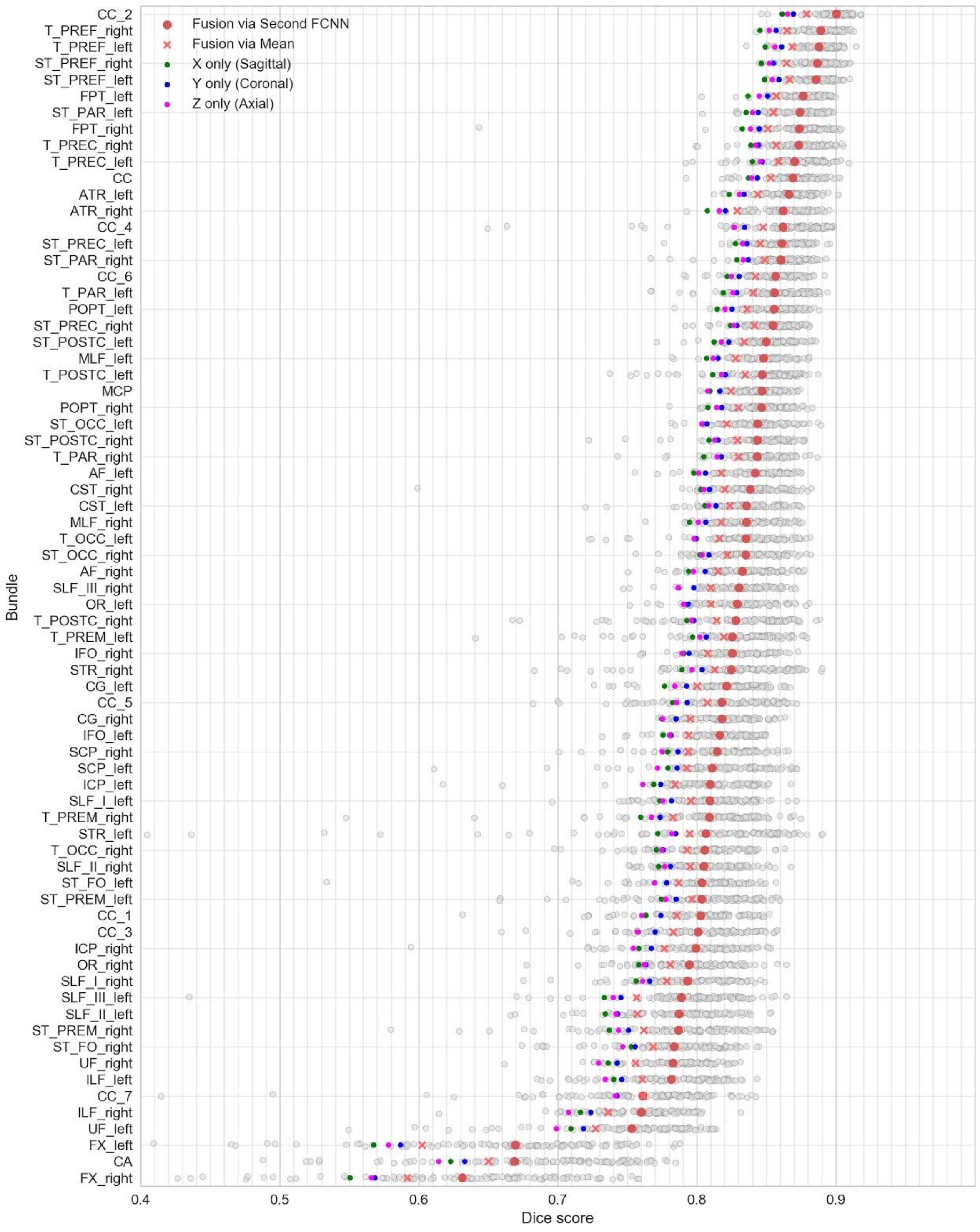

Figure 13: Mean Dice scores over all subjects for 72 tracts on the *Clinical Quality* datasets using different fusion strategies (sorted by score). Gray dots show Dice scores for each subject for the *Second FCNN* fusion strategy. The full name of each tract can be seen in figure 3.

## 3.6. Runtime

Runtime experiments were performed using a server with 48 2GHz Intel Xeon cores, or an NVIDIA Titan X for the GPU-based approaches (TRACULA and our proposed method). Table 2 shows the results for each method when segmenting all tracts in a previously unseen subject. On average, our method was 81x faster than the reference methods for *HCP Quality* and 535x faster for *Clinical Quality*.
*Multi-Mask* and *Atlas* were not included in the table since they only served as simple baseline estimation and are not openly available. Runtimes were 10.4 / 3.5 minutes (*HCP Quality* / *Clinical Quality*) and 3 / 0.4 minutes respectively.
*RecoBundles*, *TractQuerier* and *WhiteMatterAnalysis* had runtimes above 950 minutes as they depend on tractography, meaning that all the following processing steps had to be performed on thousands of streamlines. *TractQuerier* and *TRACULA* also require cortical Freesurfer parcellation, which runs for several hours.
For TractSeg, the processing step with the longest runtime was the peak extraction using the CSD, which was responsible for about 95% of the complete processing time (*HCP Quality*). The segmentation itself was quite fast, timed at about 1 minute (with *Mean* fusion only 0.5min). We employed single-shell CSD (without the multi-shell multi-tissue feature) on the *Clinical Quality* datasets, as only one shell is available here. This significantly decreased the computational burden and enabled much shorter runtimes on lower quality data for the proposed method.

Table 2: Runtime in minutes for segmenting all tracts (supported by the respective method) in a new subject using the respective method.

| | HCP Quality | Clinical Quality |
|---|---|---|
| Proposed | 20.1 | 2.2 |
| RecoBundles (Garyfallidis et al., 2017) | 1831 | 1811 |
| TractQuerier (Wassermann et al., 2016) | 1704 | 956 |
| WhiteMatterAnalysis (O`Donnell et al., 2012) | 1052 | 1042 |
| TRACULA (Yendiki et al., 2011) | 1936 | 910 |

## 3.7. Binary mask to tractogram

TractSeg only generates binary tract masks. Should streamlines be needed, these masks make it easy to generate streamlines for a certain tract. Seeding inside of the mask and only retaining the fibers which never leave the mask and end in either the cortex or brainstem, results in good tract delineations, as can be seen in figure 14. For difficult tracts like the CA, many seeds may be needed to find streamlines which cover the full length of the tract. However, as we only seed inside of the tract mask, we can use a high number of seeds without it becoming computationally infeasible.

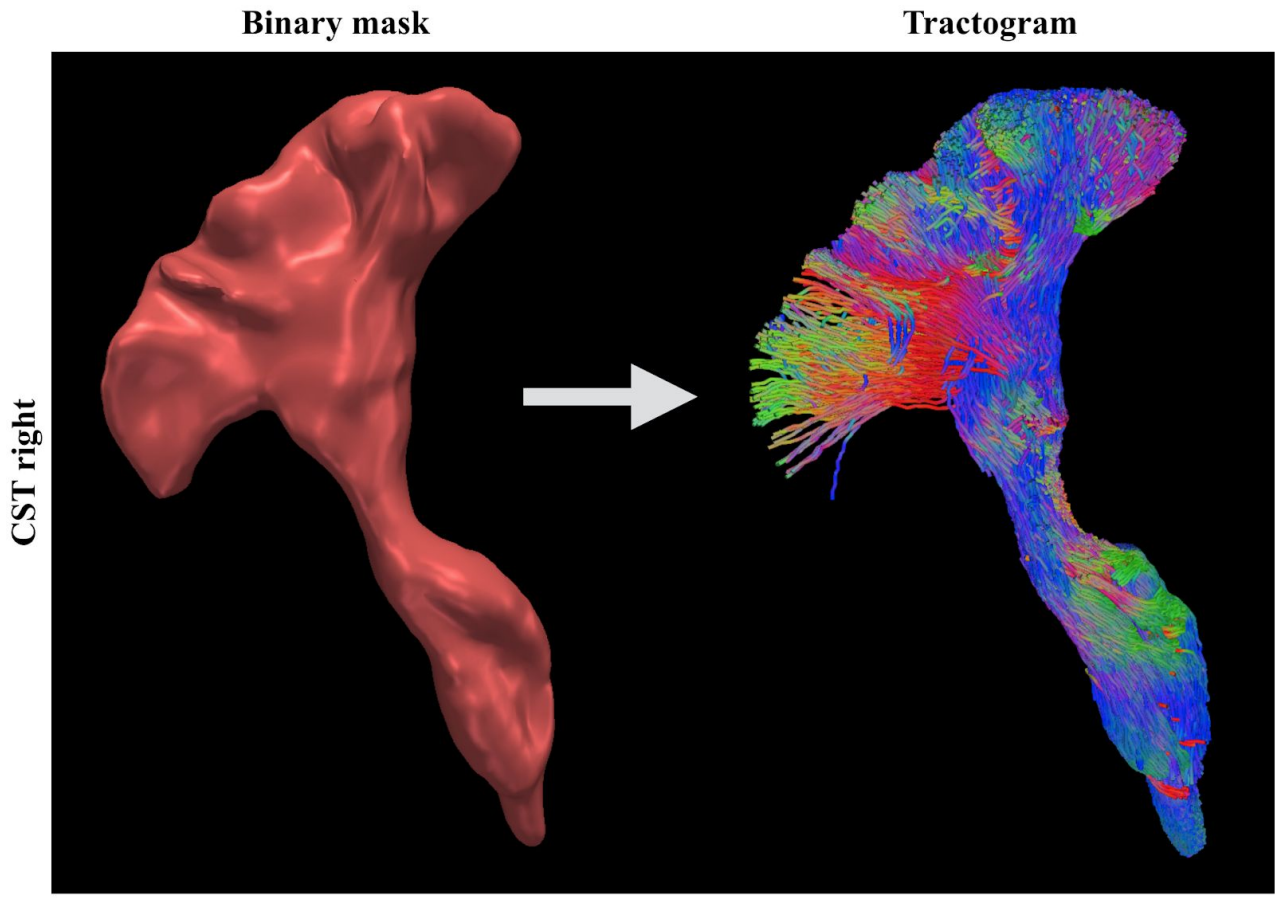

Figure 14: Using the binary mask generated by TractSeg to delineate the CST. Only fibers which never leave the mask and end in either the cortex or the brain stem were kept.

# 4. Discussion and Conclusion

## 4.1. Overview

TractSeg is a novel direct white matter tract segmentation method based on encoder-decoder FCNNs which has various advantages over the existing diffusion-weighted MR analysis pipelines. It was evaluated by segmenting a total of 72 tracts in a cohort of 105 HCP subjects in original high quality and also on reduced quality, more clinical-like datasets. Six tract segmentation methods were used as a benchmark. Our experiments demonstrate that TractSeg achieves yet unprecedented accuracy while being less affected by the reduction in resolution in the clinical quality data.

## 4.2. Reference data

Regarding the reference data, although we applied extensive efforts to mitigate the limitations of tractography, our tracts do not represent a real *ground truth*. While some remaining false positives cannot be excluded, difficult parts like the CA or the lateral projections of the CST are examples for incomplete (false negative) reconstructions in some of the subjects (see section 3.2). Moreover the reference tracts are subject to slight variations in the detailed anatomical definition of tracts, e.g. when it comes to exact start and end regions. Despite these limitations, to the best of our knowledge, the employed data set represents one of the best existing in-vivo approximations of known white matter anatomy in a cohort of that size. Moreover, as shown in section 3.2, TractSeg can deal with a certain level of inconsistency in the training data and produces accurate results in areas where the reference tracts are incomplete.

## 4.3. Reference methods

Selecting appropriate reference methods for a fair comparison was not easy as all methods have slightly different approaches and requirements. The comparison with our selected reference methods is also subject to some limitations:

*TractQuerier* was used both during the creation of the reference and during validation, inducing a potential positive bias for the method.

*WMA* evaluation was only available for 11 out of 72 tracts, which is not necessarily comparable. However, the margin between *WMA* and the proposed method remains, though, when restricting to these same 11 tracts. Moreover the mapping of clusters to anatomical tracts did not completely align with our reference tracts, as the finite set of distinct clusters offered by the atlas inherently limits a precise mapping. Due to the huge memory requirements of WMA for 10 million streamlines (>100GB) we had to restrict the analysis for *WMA* to 500k streamlines, which required approx. 30GB of RAM. This also benefited the quality of the result: cleaner clusters were created by *WMA*.

For *RecoBundles* we were only able to use 5 reference subjects due to the long runtime of *RecoBundles*. Using all 63 training subjects would have been computationally infeasible for 72 tracts and tractograms with 10 million streamlines. Moreover, as suggested by our *Atlas* and *Multi-Mask* experiments, averaging more subjects, does not necessarily increase accuracy as small details become blurred. Using 5 reference subjects therefore provides a good estimation of the performance of *RecoBundles*.

TRACULA uses its own tract definitions, which may differ from our definitions to some degree, introducing a negative bias. However, for very well defined tracts like the CST, this bias is minor and a meaningful comparison is still possible.

Our results suggested that clustering based approaches (*WMA, RecoBundles)* perform better than *TractQuerier*, which is in line with the findings of Zhang et al. (2017).

It should be noted that the straightforward *MultiMask* method produced significantly better results than the more sophisticated *RecoBundles*. Both used 5 reference subjects and were based on the reference tracts, avoiding any bias induced by different tract definitions. As *RecoBundles* works in streamline space it has to deal with the large number of false positives. Clustering helps to remove the majority of those, but not all of them. Moreover, we used the default *RecoBundles* settings. Optimising those might improve the results to some degree.

As we have shown, our comparison to the reference methods has some limitations. However, those limitations alone can not explain the large accuracy gap between our method and all reference methods, indicating that TractSeg has great potential.

## 4.4. Fusion of orientations

The high robustness of TractSeg with respect to handling different image resolution levels is likely due to the heavy data augmentation, the ensembling of multiple views and the general ability of CNNs to handle noisy data. We evaluated three different strategies to fuse the three predictions of our network for axial, sagittal and coronal input slices: Aggregation by *Mean*, aggregation by *Majority Voting* and training a *Second FCNN* to aggregate the three predictions into one final one. On the *Clinical Quality* data our *Second FCNN* approach showed a clear performance increase on the *Mean* approach, especially on the more difficult tracts (FX, CA). This confirms our assumption that a more complex fusion model can extract additional information from the different orientations. A 3D FCNN architecture would make those fusion methods obsolete, but requires more memory, prompting us to opt for a 2D architecture. Further research on 3D architecture could, however, open up promising avenues.

## 4.5. Supervised Learning

TractSeg is based on supervised learning, bearing the inherent limitation of depending on the availability and quality of training data. This is similar to *RecoBundles*, *WMA*, *Atlas*, *MultiMask* and *TRACULA* which also require reference tracts or atlases. Using scanners and acquisition sequences different from the training data introduces a domain shift, reducing performance. By using heavy data augmentation during training this domain shift can be reduced.

Our initial experiments on unseen datasets with and without pathologies are very promising and show that TractSeg is able to still produce anatomically plausible results for the vast majority of bundles. Future work could aim at adapting methods from the field of domain adaptation to further improve the results and perform more detailed evaluations on non-HCP data.

## 4.6. Complex pipelines

Most existing pipelines are based on several steps. For example TractQuerier first needs to perform streamline tracking first, then cortical parcellation and finally the actual querying. All of these steps come with their own pitfalls and inaccuracies which accumulate further down the pipeline. Moreover, all of these steps require additional runtime and installation and setup of additional tools. Our approach enables us to avoid steps such as atlas registration, tractography or parcellation, thereby circumventing many of the pitfalls introduced by those steps, resulting in higher accuracy, shorter runtime and an easier setup. A more complex pipeline and manual interaction was only required for generating the training data.

## 4.7. Code and data availability

Our dataset of 72 high quality tract dissections for 105 HCP subjects is openly available: https://doi.org/10.5281/zenodo.1088277. Up until now, the majority of papers have used their own raters to extract reference tracts. This makes meaningful comparison between methods impossible. We hope that this dataset can also help to standardize method evaluation. TractSeg is openly available as an easy-to-use python package with pretrained weights: https://github.com/MIC-DKFZ/TractSeg/

# Acknowledgments

HCP data were provided by the Human Connectome Project, WU-Minn Consortium (Principal Investigators: David Van Essen and Kamil Ugurbil; 1U54MH091657) funded by the 16 NIH Institutes and Centers that support the NIH Blueprint for Neuroscience Research; and by the McDonnell Center for Systems Neuroscience at Washington University.

Data used in preparation for this article were obtained from the SchizConnect database (http://schizconnect.org). As such, the investigators within SchizConnect contributed to the design and implementation of SchizConnect and/or provided data but did not participate in analysis or writing of this report. Data collection and sharing for this project was funded by NIMH cooperative agreement 1U01 MH097435.

BrainGluSchi data was downloaded from the COllaborative Informatics and Neuroimaging Suite Data Exchange tool (COINS;http://coins.mrn.org/dx) and data collection was funded by NIMH R01MH084898-01A1.
COBRE data was downloaded from the COllaborative Informatics and Neuroimaging Suite Data Exchange tool (COINS; http://coins.mrn.org/dx) and data collection was performed at the Mind Research Network, and funded by a Center of Biomedical Research Excellence (COBRE) grant 5P20RR021938/P20GM103472 from the NIH to Dr. Vince Calhoun.
This work was supported by the German Research Foundation (DFG) grant MA 6340/10-1 and grant MA 6340/12-1.

# Supplementary Material

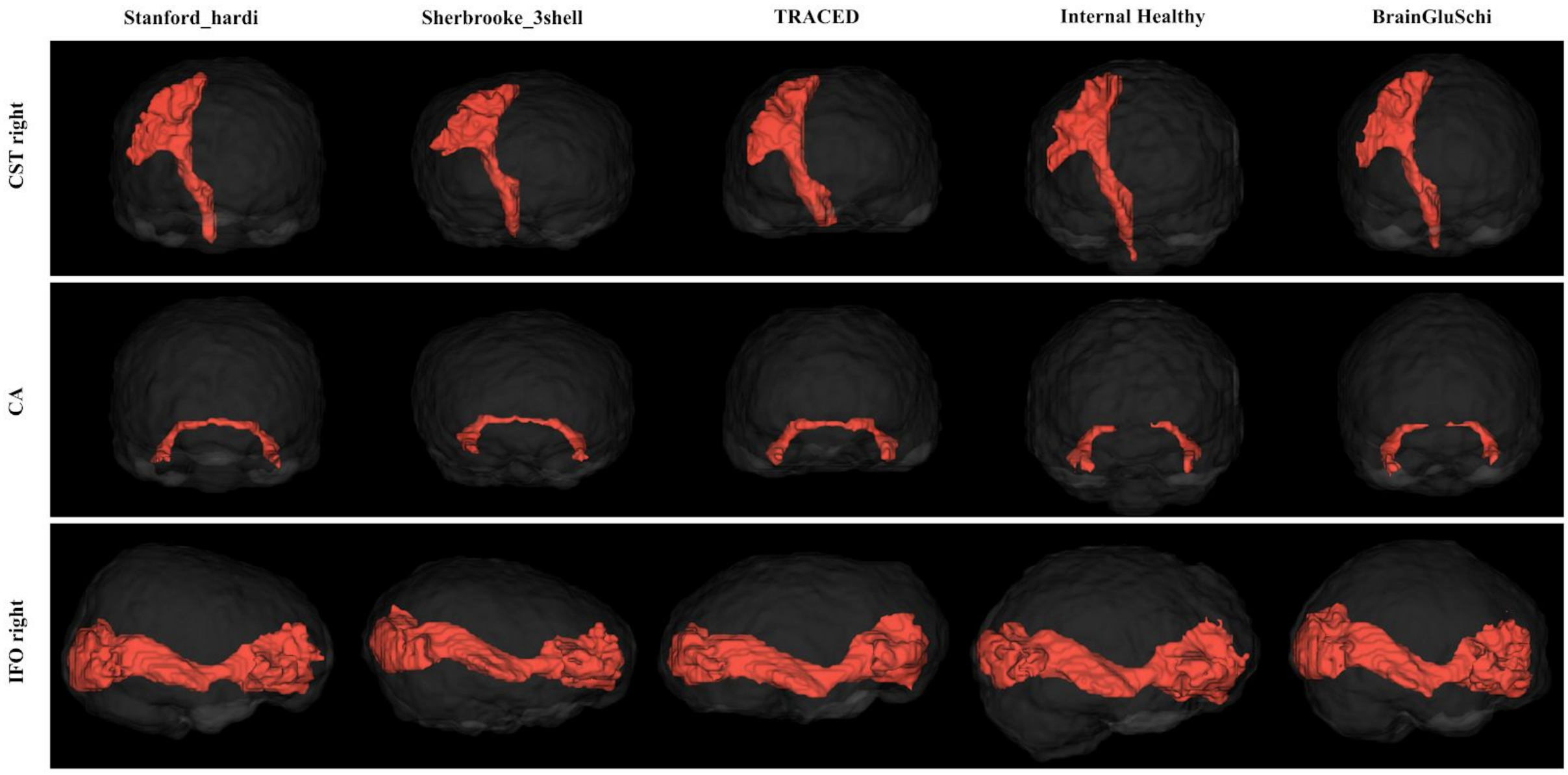

Figure 1: Qualitative results of our proposed method on 5 healthy subjects from 5 different datasets (see table 1) while being trained on the HCP dataset: reconstruction of right corticospinal tract (CST), commissure anterior (CA) and right inferior occipito-frontal fascicle (IFO).

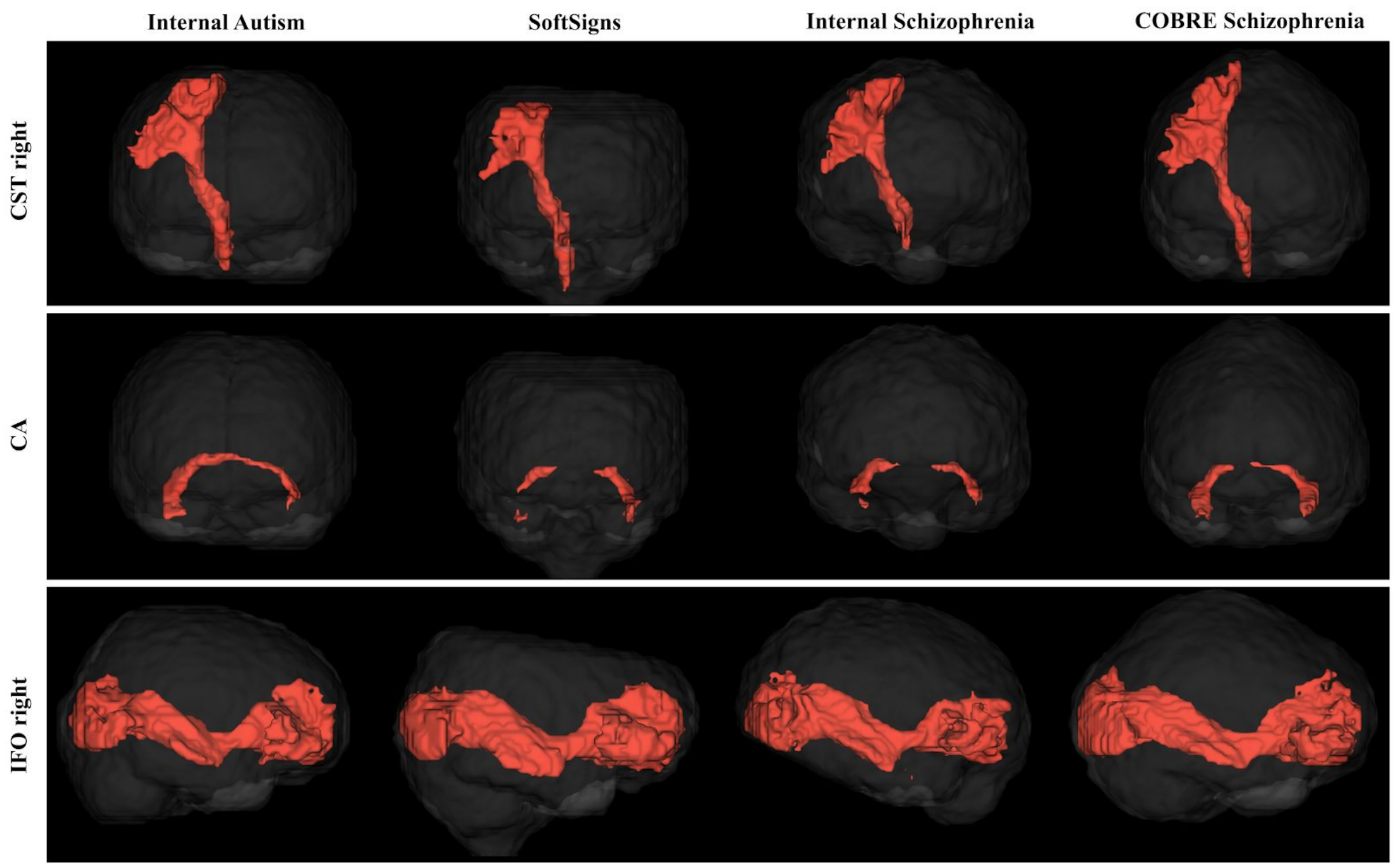

Figure 2: Qualitative results of our proposed method on 4 subjects with pathologies from 4 different datasets (see table 1) while being trained on the HCP dataset: reconstruction of right corticospinal tract (CST), commissure anterior (CA) and right inferior occipito-frontal fascicle (IFO).